\documentclass[]{spie}  

\usepackage{amsmath,amsfonts,amssymb}
\usepackage{graphicx}
\usepackage[colorlinks=true, allcolors=blue]{hyperref}
\usepackage{booktabs}
\usepackage{multirow}

\usepackage[
backend=biber,
style=numeric-comp,
sorting=none,
]{biblatex}

\title{SkillFormer: Unified Multi-View Video Understanding for Proficiency Estimation}

\author{Edoardo Bianchi}
\author{Antonio Liotta}
\affil{Free University of Bozen-Bolzano, Bozen-Bolzano, Italy}

\authorinfo{Further author information: (Send correspondence to E.B.)\\E.B.: E-mail: edbianchi@unibz.it\\  A.L.: E-mail: antonio.liotta@unibz.it}

\begin{document} 
\maketitle

\begin{abstract}
Assessing human skill levels in complex activities is a challenging problem with applications in sports, rehabilitation, and training. In this work, we present SkillFormer, a parameter-efficient architecture for unified multi-view proficiency estimation from egocentric and exocentric videos. Building on the TimeSformer backbone, SkillFormer introduces a CrossViewFusion module that fuses view-specific features using multi-head cross-attention, learnable gating, and adaptive self-calibration. We leverage Low-Rank Adaptation to fine-tune only a small subset of parameters, significantly reducing training costs. In fact, when evaluated on the EgoExo4D dataset, SkillFormer achieves state-of-the-art accuracy in multi-view settings while demonstrating remarkable computational efficiency, using 4.5x fewer trainable parameters and requiring 3.75x fewer training epochs than prior baselines. It excels in multiple structured tasks, confirming the value of multi-view integration for fine-grained skill assessment. Project page: https://edowhite.github.io/SkillFormer
\end{abstract}


\section{Introduction}
\label{sec:intro}
Automatically estimating human expertise from video is a key challenge in computer vision, especially in domains like sports, rehabilitation, and skill training. Yet, unlike action recognition, this task requires capturing subtle differences in motion quality and precision, which are often visible only through a combination of viewpoints.

To address this, we propose SkillFormer, a lightweight and efficient architecture for multi-view skill assessment. SkillFormer processes synchronized egocentric and exocentric video streams using a TimeSformer backbone, enhanced by a novel CrossViewFusion module that leverages cross-attention and learnable gating. To enable efficient fine-tuning, it incorporates Low-Rank Adaptation (LoRA). This design achieves competitive overall accuracy while improving computational efficiency through reduced trainable parameters and faster convergence.

This work introduces the following key contributions:

\begin{itemize}
\item A CrossViewFusion module that fuses multiple views via multi-head cross-attention and learnable gating, allowing the model to focus on the most informative aspects of each view.
\item A self-calibration mechanism that aligns view-specific embeddings using learnable statistics.
\item A LoRA-based fine-tuning strategy applied to the TimeSformer backbone, enabling efficient adaptation with minimal overhead.
\end{itemize}

Compared to state-of-the-art benchmarks, SkillFormer demonstrates strong overall classification accuracy in multi-view settings, which are the primary focus of our work. While our Ego-only configuration achieves comparable results to specialized single-view methods (45.9\% vs 46.8\%), the real strength of our approach emerges in multi-view scenarios. SkillFormer significantly improves accuracy in exocentric settings by 14\% (from 40.6\% to 46.3\%) and in combined ego+exo setups by 16.4\% (from 40.8\% to 47.5\%). The benefits are particularly pronounced in structured activities, where our multi-view approach achieves improvements for Basketball (77.88\%), Cooking (60.53\%), and Bouldering (33.52\%) by effectively integrating complementary spatial and temporal features from synchronized perspectives. These substantial gains highlight the effectiveness of our integrated cross-attention fusion approach, particularly in scenarios where multiple viewpoints and fine-grained motion cues are crucial for expertise assessment.

\section{Background and Related Works}
\label{sec:related}
Action Quality Assessment (AQA) focuses on evaluating human action quality with applications in sports and training \cite{AQA_survey}. Traditional approaches used handcrafted features or 3D CNNs \cite{parmar19,carreira17,c3d}, while recent methods adopt transformer architectures like TimeSformer and Video Swin for long-range dependencies \cite{timesformer,videoswin}. Specialized designs include GDLT's grade-aware features \cite{gdlt} and CoRe's contrastive regression \cite{core}.

Multi-modal AQA improves robustness by integrating complementary signals. Some approaches combine RGB and depth for industrial tasks \cite{gsfmeccano}, while others rely on skeletons to capture motion kinematics \cite{gsp, hierarchicalpose}. Datasets like EgoExo4D \cite{egoexo4d} provides multi-perspective data but baselines use separate architectures with heterogeneous pretraining, limiting unified deployment

Parameter-Efficient Adaptation through LoRA \cite{lora} enables fine-tuning with reduced parameters, addressing computational constraints in video understanding tasks. Knowledge distillation also supports low-resource adaptation for vision tasks \cite{kdohsvd}.

Our work addresses these limitations through unified multi-view architecture with efficient fusion and consistent pretraining, simultaneously improving accuracy and computational efficiency.

\begin{figure}[t]
    \centering
    \includegraphics[width=0.8\linewidth]{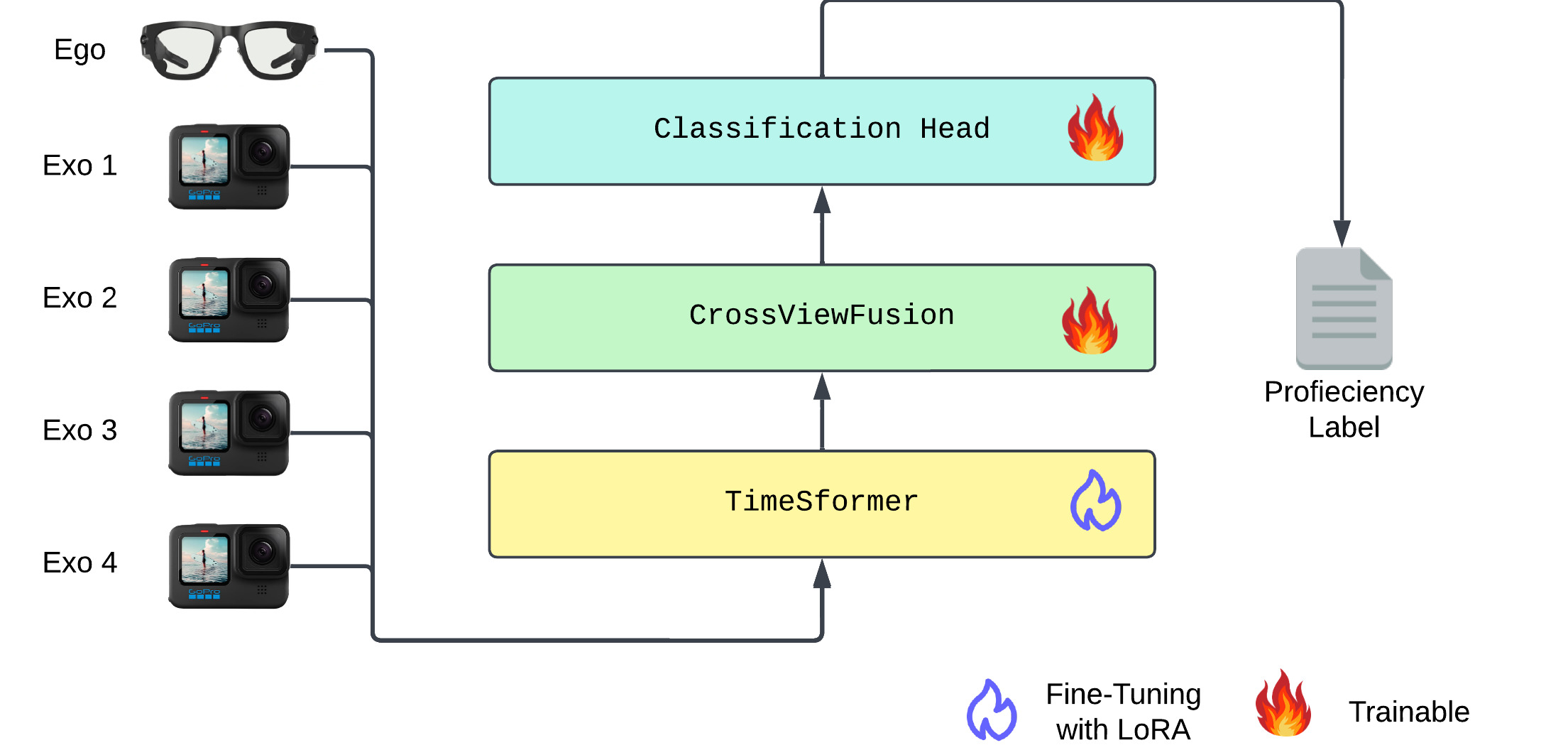}
    \caption{\label{fig:skillformer_architecture}
    Overview of the SkillFormer architecture. Multi-view video inputs (one egocentric and up to four exocentric) are processed through a shared TimeSformer backbone fine-tuned with LoRA. Features are fused using the CrossViewFusion module and passed to a classification head.}
\end{figure}

\section{Proposed Methodology}
\label{sec:methods}
We present SkillFormer, a novel architecture for unified multi-view proficiency estimation. Our model builds upon the TimeSformer \cite{timesformer} framework, incorporating parameter-efficient fine-tuning through LoRA \cite{lora} and a specialized CrossViewFusion module for multi-view integration. Figure~\ref{fig:skillformer_architecture} provides an overview of our approach.

\subsection{Problem Formulation}
\label{subsec:problem}
Given a set of synchronized videos from multiple viewpoints (e.g., first-person and third-person cameras) capturing a person performing an activity, our goal is to classify their expertise level into one of predefined categories (Novice, Early Expert, Intermediate Expert, or Late Expert). Formally, for each sample, we have a set of video sequences $\{V_1, V_2, ..., V_n\}$ from $n$ different views, and we aim to predict a label $y \in \{0, 1, 2, 3\}$ indicating the expertise level.

\subsection{Video Feature Extraction}
\label{subsec:feature_extraction}
We use a TimeSformer backbone \cite{timesformer}, pretrained on Kinetics-600 \cite{carreira17}, to extract spatio-temporal features by modeling spatial and temporal dependencies via divided attention. For efficient adaptation, we apply LoRA \cite{lora} to key components—attention projections (qkv), attention output dense layers, temporal attention components, and feed-forward layers—allowing us to fine-tune only a small number of parameters. During inference, LoRA adapters are merged with the base model weights, eliminating computational overhead while maintaining full model capacity.

For multi-view inputs, we flatten the batch and view dimensions, extract features in parallel using the shared backbone, and reshape the output before fusing it with the CrossViewFusion module. This design maintains per-view identity while enabling efficient joint processing.

To support variable-length inputs while preserving alignment with the original 8-frame pretraining, we interpolate both the input frames and the model’s temporal positional embeddings. In particular, we apply linear interpolation to the learned time embeddings, enabling the model to generalize across different temporal resolutions without degrading temporal consistency.

\begin{figure}[t]
    \centering
    \includegraphics[width=1\linewidth]{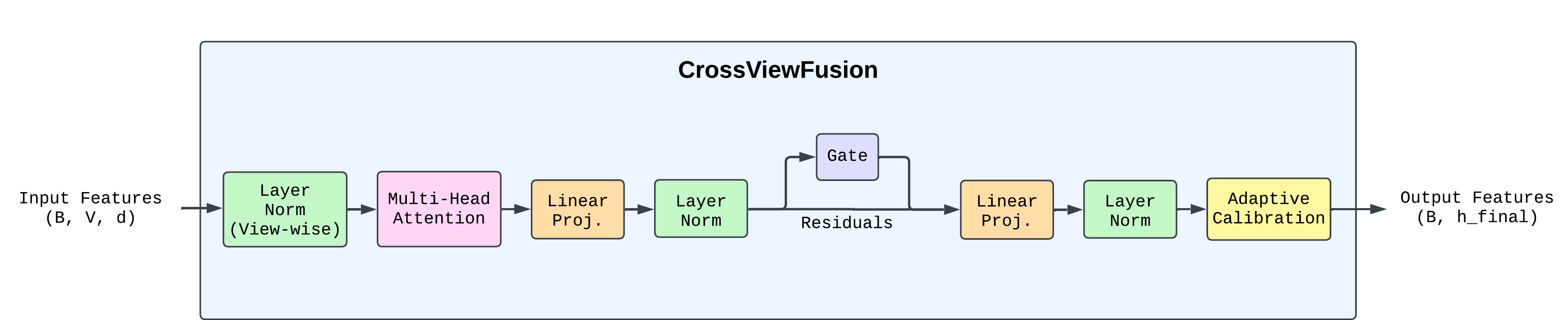}
    \caption{\label{fig:attentive_projector}
    Detailed architecture of the CrossViewFusion module. Input features $(B, V, d)$ undergo: (1) Layer normalization per view, (2) Multi-head cross-attention enabling each view to attend to all others, (3) View aggregation via mean pooling, (4) Feed-forward transformation with GELU activation, (5) Learnable gating mechanism $\mathbf{g} = \sigma(\text{Linear}(\mathbf{h}))$ for selective feature modulation, (6) Final projection, and (7) Adaptive self-calibration using learnable statistics to align with classification space.}
\end{figure}

\subsection{CrossViewFusion for Unified Multi-View Integration}
\label{subsec:attentive_projector}
We introduce CrossViewFusion, a lightweight module for fusing multi-view video streams through normalization, cross-view attention, gated transformation, and adaptive self-calibration (Figure~\ref{fig:attentive_projector}).

Given view-specific features $\mathbf{X} \in \mathbb{R}^{B \times V \times d}$ where $B$ is batch size, $V$ is number of views, and $d=768$, the module applies:

\textbf{View-wise Normalization and Multi-Head Cross-View Attention:}
To account for statistical disparities across viewpoints, we first apply layer normalization independently to each view's features. We then employ multi-head attention where each view attends to all views (including itself), enabling the model to learn inter-view dependencies and adaptively weight view-specific cues:
\begin{align}
\mathbf{X}_{norm} &= \text{LayerNorm}(\mathbf{X}) \\
\mathbf{X}_{attn}, \_ &= \text{MultiheadAttention}(Q=\mathbf{X}_{norm}, K=\mathbf{X}_{norm}, V=\mathbf{X}_{norm})
\end{align}

\textbf{View Aggregation and Feature Transformation:}
We aggregate the attention-weighted features through averaging and pass them through a feed-forward network with GELU activation, layer normalization, and dropout for regularization:
\begin{align}
\mathbf{h} &= \frac{1}{V}\sum_{v=1}^{V} \mathbf{X}_{attn}[:, v, :] \\
\mathbf{h}_{proj1} &= \text{Linear}_1(\mathbf{h}) \\
\mathbf{h}_{hidden} &= \text{Dropout}(\text{LayerNorm}(\text{GELU}(\mathbf{h}_{proj1})))
\end{align}

\textbf{Learnable Gating:}
The aggregated features are modulated through a learnable gating mechanism that produces element-wise weights via a sigmoid activation. This gate acts as a dynamic filter to selectively amplify informative features while suppressing irrelevant ones:
\begin{align}
\mathbf{g} &= \sigma(\text{Linear}_{gate}(\mathbf{h}_{hidden})) \\
\mathbf{h}_{gated} &= \mathbf{g} \odot \mathbf{h}_{hidden}
\end{align}

\textbf{Final Projection and Adaptive Self-Calibration:}
To align the fused representations with the target classification space, we apply a final linear projection followed by adaptive self-calibration. Unlike traditional fixed-statistics normalization, we learn per-feature affine parameters to reshape feature distributions in a data-driven manner:
\begin{align}
\mathbf{h}_{proj2} &= \text{Linear}_2(\mathbf{h}_{gated}) \\
\mathbf{h}_{norm} &= \text{LayerNorm}(\mathbf{h}_{proj2}) \\
\mathbf{h}_{centered} &= \mathbf{h}_{norm} - \text{mean}(\mathbf{h}_{norm}, \text{dim}=-1) \\
\mathbf{h}_{scaled} &= \frac{\mathbf{h}_{centered}}{\text{std}(\mathbf{h}_{norm}, \text{dim}=-1) + \epsilon} \\
\mathbf{h}_{final} &= \mathbf{h}_{scaled} \cdot \sigma_{learn} + \mu_{learn}
\end{align}
where $\mu_{learn}, \sigma_{learn}$ are learnable statistics that allow the model to adaptively reshape feature distributions for improved skill assessment across diverse activities.

\section{Experimental Setup}
\label{sec:experiments}

\subsection{Dataset}
\label{subsec:dataset}
We evaluate on the Ego-Exo4D dataset \cite{egoexo4d}, a large-scale benchmark with over 1,200 hours of synchronized egocentric (first-person) and exocentric (third-person) video from 740 participants across 123 real-world environments. Each sample includes one egocentric video (Project Aria glasses) and up to four time-synchronized exocentric views (static GoPro cameras).

For proficiency estimation, we use six domains (cooking, music, basketball, bouldering, soccer, dance) with four discrete levels: \textit{Novice}, \textit{Early Expert}, \textit{Intermediate Expert}, and \textit{Late Expert}. Health and bike repair domains are excluded due to imbalanced label distributions. The dataset is skewed toward intermediate and late experts due to targeted recruitment of skilled participants.

We use all videos with available proficiency estimation labels, following the official benchmark’s training and validation splits. As in prior work \cite{egoppg}, we train on the official training set, reserving 10\% for validation, and evaluate on the full, held-out official validation set. This setup ensures direct and fair comparability with the accuracy reported in the dataset paper.

\subsection{Implementation Details}
\label{subsec:implementation}
Models use a TimeSformer backbone \cite{timesformer} pretrained on Kinetics-600 \cite{carreira17}, fine-tuned for 4 epochs with AdamW (weight decay=0.01), cosine annealing schedule, batch size 16, on single A100 GPU. 

Videos are center-cropped to $224 \times 224$, rescaled to the $[0,1]$ range, and normalized with mean $0.45$ and standard deviation $0.225$. Frames are sampled uniformly with temporal interpolation of positional embeddings for variable lengths. 

Hyperparameters such as the number of input frames, projector hidden size, LoRA rank and scaling, and learning rate were tuned based on the number and type of input views. All other settings were held fixed across experiments (Refer to Section \ref{sec:efficiency} and Table \ref{tab:training_config}).

\begin{table}[t]
\centering
\caption{Comparison with EgoExo4D proficiency estimation baselines \cite{egoexo4d} and with EgoPulseFormer \cite{egoppg}, a concurrent approach. We report accuracy (\%) for egocentric (Ego), exocentric (Exos), and combined views (Ego+Exos). SkillFormer outperforms the baselines in the Exos and Ego+Exos settings with significantly fewer trainable parameters and fewer epochs. Bold denotes the best accuracy; underlined values indicate second-best.\\}
\begin{tabular}{lcccccc}
\toprule
\textbf{Method} & \textbf{Pretrain} & \textbf{Ego} & \textbf{Exos} & \textbf{Ego+Exos} & \textbf{Params} & \textbf{Epochs} \\
\midrule
Random & - & 24.9 & 24.9 & 24.9 & - & - \\
Majority-class & - & 31.1 & 31.1 & 31.1 & - & - \\
TimeSformer & - & 42.3 & 40.1 & 40.8 & 121M & 15 \\
TimeSformer & K400 & 42.9 & 39.1 & 38.6 & 121M & 15 \\
TimeSformer & HowTo100M & \textbf{46.8} & 38.2 & 39.7 & 121M & 15 \\
TimeSformer & EgoVLP & 44.4 & \underline{40.6} & 39.5 & 121M & 15 \\
TimeSformer & EgoVLPv2 & \underline{45.9} & 38.0 & 37.8 & 121M & 15 \\
\midrule
EgoPulseFormer & EgoPPG-DB & 45.3 & 35.9 & \underline{42.4} & 121M & 15 \\
\midrule
\textbf{SkillFormer-Ego} & K600 & \underline{45.9} & - & - & 14M & 4 \\
\textbf{SkillFormer-Exos} & K600 & - & \textbf{46.3} & - & 20M & 4 \\
\textbf{SkillFormer-EgoExos} & K600 & - & - & \textbf{47.5} & 27M & 4 \\
\bottomrule
\end{tabular}
\label{tab:baseline_comparison}
\end{table}

\section{Results}
\label{subsec:results}
We evaluate SkillFormer on the EgoExo4D \cite{egoexo4d} proficiency estimation benchmark using classification accuracy as the primary evaluation metric, in line with the official proficiency demonstrator benchmark protocol. We consider three input configurations: egocentric (Ego, 1 view), exocentric (Exos, 4 views), and multi-view (Ego+Exos, 5 views), enabling us to isolate the contribution of each modality (Sections~\ref{sec:overall} and~\ref{sec:efficiency}). Additionally, we analyze per-domain accuracy to evaluate generalization across diverse real-world tasks (Section~\ref{sec:perdomain}). SkillFormer achieves consistent improvements in both accuracy and efficiency, outperforming multi-view baselines with fewer trainable parameters and significantly lower training cost.

\subsection{Overall Accuracy and Efficiency}
\label{sec:overall}
Table~\ref{tab:baseline_comparison} compares overall accuracy between SkillFormer and baseline models from the EgoExo4D benchmark~\cite{egoexo4d}. In contrast to baselines—which train separate models for egocentric and exocentric inputs and perform late fusion at inference—SkillFormer uses a single unified model for each configuration (Ego, Exos, Ego+Exos). Furthermore, we do not apply multi-crop testing, simplifying inference and reducing computational overhead.

SkillFormer achieves state-of-the-art classification accuracy in both Exos (46.3\%) and Ego+Exos (47.5\%) settings, outperforming the best TimeSformer baseline by up to 16.4\%. Simultaneously, SkillFormer demonstrates increased computational efficiency by using 4.5x fewer trainable parameters (27M vs. 121M) and requiring 3.75x fewer training epochs (4 vs. 15). These results highlight SkillFormer’s accuracy and compute-efficiency. Random and majority-class baselines perform significantly worse, underscoring the inherent complexity of the task.

It is worth noting that the proficiency label distribution is notably imbalanced, skewed toward intermediate and late experts due to targeted recruitment of skilled participants. This may bias overall accuracy by underrepresenting novice classes.

\subsection{Scaling Strategy and Parameter Analysis}
\label{sec:efficiency}
Table~\ref{tab:training_config} details our scaling strategy across view configurations. As the number of views increases, our design prioritizes efficiency without compromising accuracy. To this end, we strategically reduce the number of frames per view (32→16)—preserving the temporal span with fewer sampled tokens—while proportionally increasing the LoRA rank (32→64), alpha (64→128), and hidden dimension (1536→2560). This trade-off compensates for reduced per-view tokens by enabling richer cross-view transformations.

Our choice of LoRA rank and fusion dimensionality is not arbitrary. Higher ranks allow the adapter to express richer transformations across views, compensating for the reduced token budget. Empirically, we found that increasing these parameters moderately yields significant gains in accuracy while maintaining training efficiency within tractable computational budgets. For example, SkillFormer-Ego+Exos achieves 47.5\% accuracy with only 27M trainable parameters—a 4.5x reduction compared to full fine-tuning of the 121M parameter TimeSformer backbone—demonstrating the effectiveness of low-rank adaptation and targeted fusion.

This design reflects a key motivation behind SkillFormer: enabling scalable, parameter-efficient skill recognition across multi-view egocentric and exocentric inputs.

\begin{table}[t]
\centering
\caption{SkillFormer scaling strategy across view configurations. As views increase, we reduce frames while proportionally increasing LoRA rank and fusion capacity to maintain model expressiveness within computational constraints. Fixed across all runs: epochs = 4, batch size = 16, output dim = 768, attention heads = 16. \textit{Hid} denotes the dimensionality of the intermediate representation used in the CrossViewFusion module.\\}
\begin{tabular}{lccccccc}
\toprule
\textbf{Views} & \textbf{Frames} & \textbf{LoRA-r} & \textbf{LoRA-a} & \textbf{Hid} & \textbf{LR} & \textbf{Params} & \textbf{Acc} (\%) \\
\midrule
Ego & 32 & 32 & 64 & 1536 & 5e-5 & 14M & 45.9 \\
Exos & 24 & 48 & 96 & 2048 & 3e-5 & 20M & 46.3 \\
Ego+Exos & 16 & 64 & 128 & 2560 & 2e-5 & 27M & 47.5 \\
\bottomrule
\end{tabular}
\label{tab:training_config}
\end{table}

\begin{table}[t]
\centering
\caption{Per-scenario accuracy (\%) for the majority-class baseline, the baseline models \cite{egoexo4d}, and SkillFormer across different view configurations. Bold indicates the best-performing method per scenario, while underlined values represent the second-best. SkillFormer consistently outperforms other models in multi-view setups, particularly in Basketball, Cooking, and Bouldering.\\}
\begin{tabular}{lccccccc}
\toprule
\multirow{2}{*}{\textbf{Scenario}} & \multirow{2}{*}{\textbf{Majority}} & \multicolumn{3}{c}{\textbf{Baseline}} & \multicolumn{3}{c}{\textbf{SkillFormer}} \\
\cmidrule(lr){3-5} \cmidrule(lr){6-8}
& & Ego & Exos & Ego+Exos & Ego & Exos & Ego+Exos \\
\midrule
Basketball & 36.19 & 51.43 & 52.30 & 55.24 & 69.03 & \underline{70.80} & \textbf{77.88} \\
Cooking    & \underline{50.00} & 45.00 & 35.00 & 35.00 & 31.58 & 47.37 & \textbf{60.53} \\
Dancing    & \underline{51.61} & \textbf{55.65} & 42.74 & 42.74 & 20.51 & 15.38 & 13.68 \\
Music      & 58.97 & 46.15 & \underline{69.23} & 56.41 & \textbf{72.41} & 68.97 & 68.10 \\
Bouldering & 0.00 & 25.31 & 17.28 & 17.28 & 30.77 & \textbf{33.52} & \underline{31.87} \\
Soccer     & 62.50 & 56.25 & \textbf{75.00} & \textbf{75.00} & \underline{70.83} & 66.67 & 66.67 \\
\bottomrule
\end{tabular}
\label{tab:per_scenario_accuracy}
\end{table}

\subsection{Per-Domain Accuracy Analysis}
\label{sec:perdomain}
Table~\ref{tab:per_scenario_accuracy} reports per-scenario accuracy across all models and configurations. SkillFormer consistently outperforms baselines in the Ego+Exos setting for structured and physically grounded activities such as \textit{Basketball} (77.88\%) and \textit{Cooking} (60.53\%). These domains benefit from synchronized egocentric and exocentric perspectives, which enable better modeling of spatial layouts and temporally extended actions. The fusion of multi-view signals allows SkillFormer to exploit cross-perspective cues, such as object-hand interactions or full-body movement trajectories, which are often ambiguous or occluded in single-view inputs.

Interestingly, in \textit{Music}, the Ego-only configuration achieves the highest accuracy (72.41\%), suggesting that head-mounted views are sufficient to capture detailed instrument manipulation, while additional views may introduce redundant or misaligned information. Similarly, in \textit{Bouldering}, the Exos-only configuration outperforms both Ego (30.77\%) and Ego+Exos (31.87\%) with 33.52\% accuracy, indicating that third-person perspectives better capture full-body spatial positioning and climbing technique assessment. This highlights the flexibility of SkillFormer to adapt its fusion mechanism to view-specific signal quality, but also demonstrates that domain-specific viewpoint advantages can be diluted when combining perspectives that provide conflicting or redundant information.

Subjective domains like \textit{Dancing} reveal limitations: SkillFormer’s Ego+Exos accuracy (13.68\%) falls significantly below both the majority baseline (51.61\%) and the baseline Ego model (55.65\%). This indicates that tasks with high intra-class variability and weak structure may not benefit as much from multi-view fusion, or may require additional modalities (e.g., audio) or supervision cues to disambiguate subtle skill indicators.

These trends underscore that SkillFormer is particularly effective in domains requiring precise spatial-temporal reasoning and multi-view integration, aligning with its architectural design focused on efficient yet expressive fusion across diverse camera perspectives.

However, this per-domain evaluation is also affected by class imbalance: domains with few novice samples may exhibit inflated or unstable accuracy due to limited representation of early-stage performers.

\section{Limitations and Future Work}
\label{sec:limitations}
Despite strong accuracy and efficiency, SkillFormer has several limitations suggesting future improvements.

Fixed frame sampling may miss critical temporal cues in variable activities, a limitation that could be addressed through adaptive sampling with attention-based keyframe detection or sport-tailored sampling methods \cite{spotting}. SkillFormer also underperforms significantly in subjective domains like Dancing, suggesting challenges with stylistic variability that domain adaptation strategies such as curriculum learning could improve.

Additionally, the class imbalance in EgoExo4D affects evaluation accuracy, which could be mitigated through class-weighted loss functions or synthetic data generation. Finally, SkillFormer lacks semantic understanding of skill components, limiting interpretability—a challenge that could be addressed through multi-task learning with pose estimation or text-guided supervision. These enhancements would improve SkillFormer's accuracy in challenging domains while maintaining its computational efficiency.

\section{Conclusions}
\label{sec:conclusion}
We introduced SkillFormer, a lightweight architecture for multi-view skill assessment that leverages synchronized egocentric and exocentric video inputs. Our approach combines a TimeSformer backbone with a novel CrossViewFusion module, featuring cross-attention, learnable gating, and parameter-efficient LoRA fine-tuning.

SkillFormer makes a dual contribution to multimodal video understanding. First, it achieves state-of-the-art overall classification accuracy on the EgoExo4D proficiency estimation benchmark—particularly in Exos (46.3\%) and Ego+Exos (47.5\%) settings—outperforming previous baselines by up to 16.4\%. Second, it demonstrates strong computational efficiency, using 4.5x fewer trainable parameters and requiring 3.75x fewer training epochs compared to prior approaches.

Our per-domain analysis reveals that SkillFormer excels in structured physical activities that benefit from complementary viewpoints, achieving significantly higher accuracy in Basketball (77.88\%), Cooking (60.53\%), and Bouldering (33.52\%). These improvements are most pronounced when both egocentric and exocentric perspectives provide complementary information about spatial layouts and temporally extended actions.

These results demonstrate that parameter-efficient adaptation and strategic multi-view fusion can substantially enhance skill assessment capabilities while reducing computational demands. SkillFormer's unified architecture for integrating egocentric and exocentric signals presents a significant advancement in efficient multi-view skill assessment, with particular promise for resource-constrained deployment scenarios.

\section*{Acknowledgements}
We acknowledge ISCRA for awarding this project access to the LEONARDO supercomputer, owned by the EuroHPC Joint Undertaking, hosted by CINECA (Italy).

\printbibliography[title={REFERENCES}]



\end{document}